# SPADE: A Large Language Model Framework for Soil Moisture Pattern Recognition and Anomaly Detection in Precision Agriculture

Yeonju Lee[a], Rui Qi Chen[a], Joseph Oboamah[b,c], Po Nien Su[d], Wei-zhen Liang[b,d], Yeyin Shi[d], Lu Gan[e], Yongsheng Chen[f], Xin Qiao[b,d], Jing Li*[a]

[a] *H. Milton Stewart School of Industrial and Systems Engineering, Georgia Institute of Technology, Atlanta, GA, USA*
[b] *Panhandle Research and Extension Center, University of Nebraska-Lincoln, Scottsbluff, NE, USA*
[c] *Department of Computer Science and Engineering, University of Nebraska-Lincoln, Lincoln, NE, USA*
[d] *Department of Biological Systems Engineering, University of Nebraska-Lincoln, Lincoln, NE, USA*
[e] *Institute for Robotics and Intelligent Machines, Georgia Institute of Technology, Atlanta, GA, USA*
[f] *School of Civil & Environmental Engineering, Georgia Institute of Technology, Georgia Institute of Technology, Atlanta, GA, USA*

**Abstract**

Accurate interpretation of soil moisture patterns is critical for irrigation scheduling and crop management, yet existing approaches for soil moisture time-series analysis either rely on threshold-based rules or data-hungry machine learning or deep learning models that are limited in adaptability and interpretability. In this study, we introduce SPADE (Soil moisture Pattern and Anomaly DEtection), an integrated framework that leverages large language models (LLMs) to jointly detect irrigation patterns and anomalies in soil moisture time-series data. SPADE utilizes ChatGPT-4.1 for its advanced reasoning and instruction-following capabilities, enabling zero-shot analysis without requiring task-specific annotation or fine-tuning. By converting time-series data into a textual representation and designing domain-informed prompt templates, SPADE identifies irrigation events, estimates net irrigation gains, detects, classifies anomalies, and produces structured, interpretable reports. Experiments were conducted on real-world soil moisture sensor data from commercial and experimental farms cultivating multiple crops across the United States. Results demonstrate that SPADE outperforms the existing method in anomaly detection, achieving higher recall and F1 scores and accurately classifying anomaly types. Furthermore, SPADE achieved high precision and recall in detecting irrigation events, indicating its strong capability to capture irrigation patterns accurately. SPADE's reports provide interpretability and usability of soil moisture analytics. This study highlights the potential of LLMs as scalable, adaptable tools for precision agriculture, which is capable of integrating qualitative knowledge and data-driven reasoning to produce actionable insights for accurate soil moisture monitoring and improved irrigation scheduling from soil moisture time-series data.

## 1. Introduction

Global crop production systems are facing mounting challenges due to climate change, population growth, and water scarcity (Farooq et al., 2023). These challenges demand more resource-efficient agricultural strategies. Irrigated agriculture contributes substantially to global crop yields, which accounts for the majority of freshwater use. Improving irrigation practices is widely recognized as a key approach to enhance water use efficiency (Muleke et al., 2023). In this context, the ability to monitor and accurately interpret soil moisture dynamics has become essential for guiding data-informed irrigation decisions (Mittelbach et al., 2012). The widespread deployment of soil moisture sensors across agricultural fields has enabled the collection of high-frequency time-series data, facilitating data-driven decision-making in precision agriculture (Bandaru et al., 2024).

Timely and accurate interpretation of soil moisture patterns in time-series data offers actionable insights for improving irrigation (Jalilvand et al., 2019). Identifying soil moisture patterns helps farmers optimize irrigation scheduling, improve water use efficiency, and increase crop yields. Under typical irrigation or rainfall conditions, soil moisture exhibits a characteristic temporal pattern, with a sharp rise following an irrigation event and a subsequent gradual decline due to infiltration and evapotranspiration as illustrated in the leftmost panel of Figure 1 (Bandaru et al., 2024). However, as shown in the middle and rightmost panels, normal patterns can exhibit considerable variability due to environmental fluctuations such as precipitation, evapotranspiration, soil heterogeneity, and crop-specific differences (Dorigo et al., 2013). In other words, defining a clear and consistent notion of normal irrigation patterns remains difficult.

At the same time, real-world sensor data often contain anomalies caused by faulty installations, hardware issues, or environmental noise (Bandaru et al., 2024). These anomalies hinder accurate interpretation of the soil moisture patterns and could lead to misinformed irrigation decisions. This underscores the importance of detecting anomalies in soil moisture data, while the variability in normal irrigation patterns makes it challenging to distinguish anomalies from normal patterns. Researchers have explored a range of approaches, from traditional statistical methods to deep learning, to detect anomalies in soil moisture time-series (Dorigo et al., 2013; Catalano et al., 2022; Bandaru et al., 2024). Dorigo et al. proposed an anomaly detection method flagging soil moisture measurements that fail to meet specified criteria, including both threshold-based absolute values and pattern-based spectral characteristics using derivatives (Dorigo et al., 2013). Recently, deep learning (DL) has been adopted to detect anomalies in soil moisture time-series (Catalano et al., 2022; Bandaru et al., 2024). Catalano et al. trained multiple linear regression and Long Short-Term Memory (LSTM) models to predict soil moisture measurements, then identified anomalies by thresholding the difference between predicted and true values (Catalano et al., 2022). Bandaru et al. also proposed an LSTM model for anomaly detection (Bandaru et al., 2024).

*Corresponding author. Email: jli3175@gatech.edu

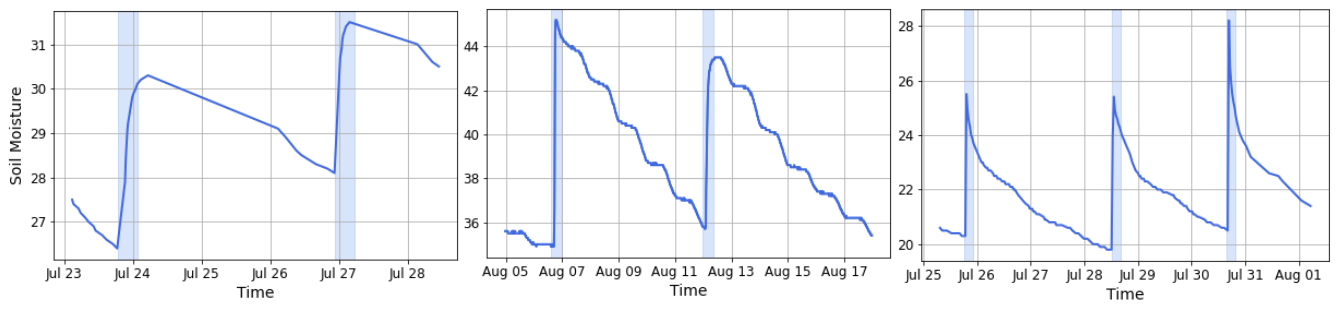

Figure 1: Temporal patterns of soil moisture under normal irrigation and rainfall conditions. The x-axis represents time, and the y-axis represents volumetric soil moisture content (%). All three plots illustrate normal irrigation patterns, with irrigation events highlighted in light blue. The leftmost figure depicts an irrigation/rainfall event characterized by a sharp increase followed by a gradual decline in moisture. The middle figure shows a step-wise decline after each sharp increase. The rightmost figure shows a pattern with a sharp increase followed by a relatively sharp decline, distinguishing it from the more gradual (leftmost) and step-wise (middle) drying patterns.

Despite these advances, key limitations remain. Most existing methods require a large amount of labeled training data to distinguish anomalies from normal patterns. This results in considerable costs for data collection and annotation. Threshold-based methods do not depend on training data, but their performance is sensitive to manually defined thresholds and thus lacks robustness across varying field conditions. In addition, these approaches primarily focus on detecting anomalies without simultaneously detecting normal irrigation patterns, which are crucial for irrigation planning and understanding crop water use. Furthermore, existing methods lack of interpretability, as they rarely provide explanations for why a particular event is identified as an anomaly, making the results difficult for farmers to understand and act upon.

Recent developments in large language models (LLMs) present an alternative paradigm for time-series analysis. LLMs are pre-trained on large and diverse textual corpora and demonstrate strong zero-shot capabilities that enable them to perform tasks without task-specific labeled data or fine-tuning(Liu et al., 2024b,a; Russell-Gilbert et al., 2024; Alnegheimish et al., 2024). This pre-training allows LLMs to integrate domain-specific instructions, reason over complex data, and produce natural language outputs that explicitly articulate their reasoning. Importantly, they can leverage prior knowledge from other domains in which they were trained, enabling transfer of reasoning strategies across contexts. These characteristics make LLMs particularly attractive for applications that require flexibility, scalability, and interpretability in analyzing high-variability time-series data such as soil moisture.

In this study, we present an integrated framework, SPADE (Soil moisture Pattern and Anomaly DEtection), which identifies irrigation patterns and detects anomalies in volumetric soil moisture data using a LLM. For our experiments, we employ ChatGPT-4.1, a LLM developed and released by OpenAI in April 2025 (Achiam et al., 2023; OpenAI, 2025). This model was selected for its advanced reasoning and instruction-following capabilities, which are critical for interpreting complex temporal soil moisture patterns and producing consistent, structured reports (OpenAI, 2025). Given weekly volumetric soil moisture time-series collected from actual farms growing multiple crops, SPADE generates a structured report that includes: (1) identification and timing of irrigation events; (2) estimation of irrigation amounts; (3) detection and classification of anomalies; and (4) descriptive and interpretable summaries for each anomaly event. Furthermore, this study examines how individual prompt instructions affect LLM behavior through representative examples. We provide insights on designing prompts that enable LLM to detect irrigation and anomalies, which can contribute to more reliable and accurate applications of LLMs in precision agriculture.

The main contributions of this study are as follows:

- To the best of our knowledge, SPADE is the first framework to leverage a LLM for generating structured and interpretable analyses of soil moisture time-series data without additional model training. We introduce a tailored prompt design that enables ChatGPT-4.1 to handle variability in soil moisture dynamics and to distinguish irrigation events from anomalies effectively.

- SPADE introduces an integrated framework that simultaneously identifies irrigation patterns and detects anomalies, addressing the limitations of existing methods and enabling a more comprehensive understanding of soil moisture dynamics.

- We validate SPADE on a diverse collection of real-world soil moisture data from farms cultivating diverse crop types. The result demonstrates its effectiveness in identifying irrigation events, detecting anomalies, and producing structured reports that support data-driven decision-making in precision agriculture.

## 2. Materials and methods

### *2.1. Data collection and preparation*

In this study, we collected time-series, volumetric soil moisture content in percentage points from 47 farm-fields growing multiple crop types (corn, sugar beet, dry bean, grapes) during 2022, 2023, 2024. Each farm-field was equipped with a 90 cm long Sentek Drill and Drop soil moisture probe (Sentek Sensor Technologies, Australia) that measures soil moisture at 10 cm increments. The duration of the time series varied across fields depending on the crop growth period, most commonly spanning about three months (July to October), although in some cases it extended up to ten months. In general, soil moisture content values range between 0% and 60%. While some farms reported soil moisture values at fixed 15-minute intervals, others exhibited irregular sampling intervals due to variations in data logging configurations.

For this study, we extracted weekly portions of the soil moisture time-series data at individual depths and constructed 100 univariate time series. The extraction of weekly portions was adopted to make the data manageable for analysis under resource constraints, which are described later in this section. Multiple depths were randomly selected from each farm to assess the model's ability to detect irrigation patterns and anomalies at different depths, resulting in separate univariate time series for each farm. We identified eight types of anomalies in the dataset: Single Spike (an isolated sharp increase), Single Dip (an isolated sharp decrease), Multiple Spikes (repeated sharp increases), Persistent Level Shift Up/Down (a sustained upward or downward shift in soil moisture levels), Transient Level Shift Up/Down (a temporary upward or downward shift that returns to baseline), and Missing Value (data gaps due to sensor or transmission errors). When extracting weekly portions, we deliberately selected some to include both anomalies and diverse irrigation patterns, while the remaining portions were randomly chosen. Most portions covered 7 days, although longer portions were used when necessary to capture Missing Value anomalies.

To enable precise identification of both irrigation events and anomalous patterns, the soil moisture data was organized into a tabular format, where each row represents an observation consisting of a soil moisture value indexed by a corresponding timestamp (YYYY-MM-DD HH:MM:SS). This tabular data was then converted into a text format by serializing each row as a line of text, preserving the chronological order and structure. This textual representation was designed to align with the input expectations of LLMs, allowing them to reason over and recongize temporal patterns without requiring additional preprocessing or encoding (Liu et al., 2024b; Russell-Gilbert et al., 2024).

Unlike existing studies (Liu et al., 2024b; Alnegheimish et al., 2024), we did not apply scaling to the soil moisture values. This decision was based on the importance of preserving the absolute magnitude of the measurements, as values outside the normal range (e.g., 0–60%) may indicate abnormal conditions. Scaling could obscure these thresholds and compromise anomaly detection. Furthermore, soil moisture value was rounded to the first decimal place, avoiding redundancy caused by excessive decimal precision and allowing values to be represented as compact tokens.

LLMs have constraints on both token window size and tokens-per-minute (TPM) throughput. We selected ChatGPT-4.1 for its strong zero-shot capabilities and consistent performance in language-based reasoning, which aligned well with the needs of our task, despite the cost. This model supports a maximum of 30,000 TPM and a context window of 100,000 tokens (OpenAI, 2025). To comply with these constraints, we extracted weekly portions of the time series. This approach ensured that each portion remained within the token and context window limits of the LLM. Although this design choice was initially driven by the computational constraints of LLMs, it also aligns with practical irrigation cycles observed in real-world farming operations.

### 2.2. *LLM-based prompt engineering for soil moisture pattern detection*

We introduce SPADE, a framework that leverages LLM to analyze soil moisture time-series data and generate structured reports on irrigation patterns and anomalies based on domain-specific instructions. Figure 2 illustrates the overall workflow of SPADE. A carefully designed prompt and soil moisture time-series data are provided as inputs, and SPADE returns a structured report summarizing irrigation patterns and anomalies. SPADE offers four main advantages over conventional time-series anomaly detection methods. First, it leverages the zero-shot reasoning capabilities of ChatGPT 4.1 to analyze soil moisture time-series data without requiring any training data, thereby substantially reducing the burden of data collection and annotation. Second, unlike most existing methods that only focus on anomaly detection, SPADE is also capable of identifying normal irrigation and rainfall patterns. This helps farmers make informed irrigation decisions. Third, SPADE enhances interpretability by generating structured textual reports including human-readable explanations of each detected anomaly and irrigation event. Fourth, SPADE is scalable, allowing it to incorporate new types of anomalies and adapt to changing irrigation patterns.

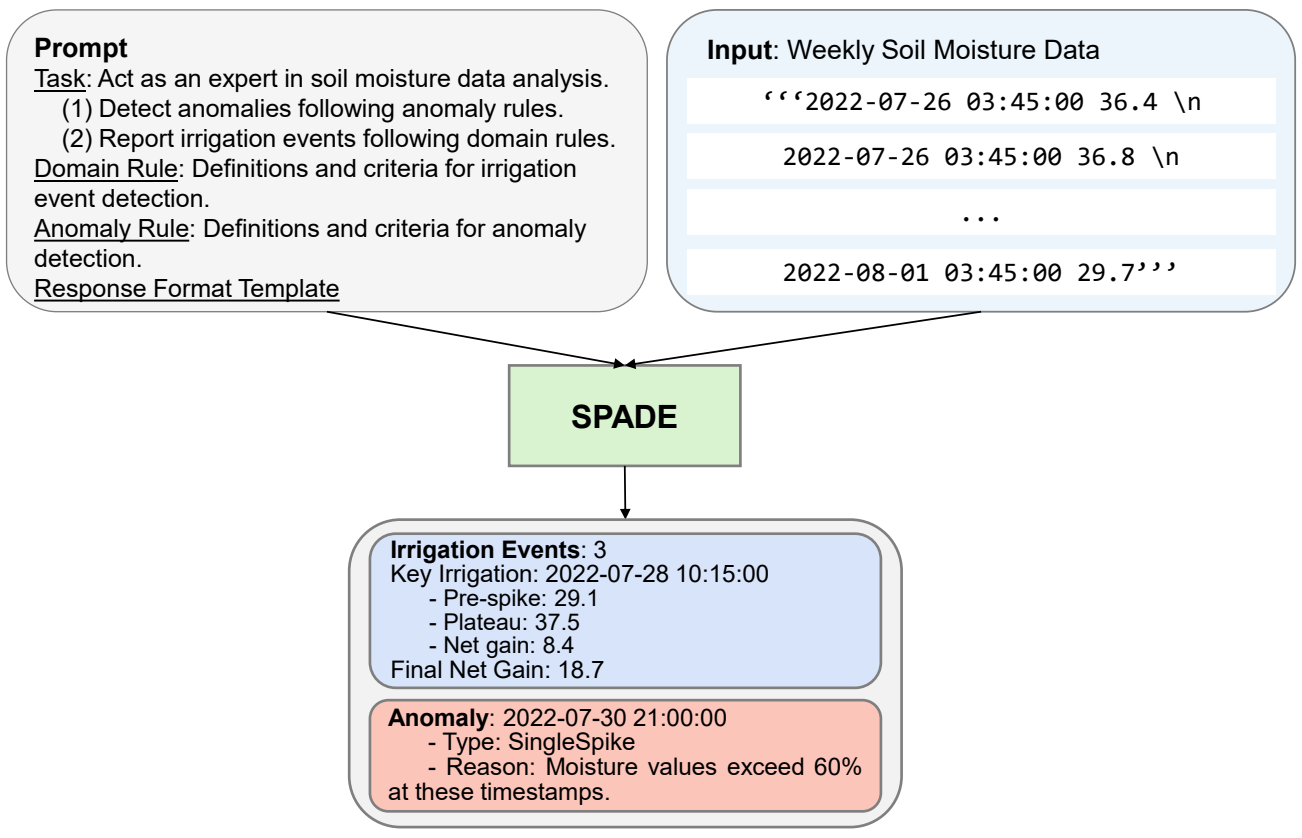

Figure 2: SPADE framework for structured soil moisture analysis. A carefully designed prompt guides SPADE to analyze weekly volumetric soil moisture data, identify irrigation events, and detect anomalies using domain and anomaly rules, while following a response template. Based on the prompt and data, SPADE generates a structured report that provides insights into irrigation events and detected anomalies. The prompt shown here is a brief overview. A more detailed version is provided later in the paper along with additional report examples.

To achieve structured and interpretable analysis, SPADE is guided by a prompt template rather than a trained model. As shown in Figure 2, this template defines (i) the task by assigning the LLM the role of an expert in agricultural water management, (ii) the input format so that time-series data are interpreted consistently, (iii) a set of anomaly rules that specify how anomaly should be identified and classified, (iv) a set of domain rules that describe normal irrigation and rainfall events, and (v) the response format that dictates how the findings should be summarized in a standardized structure. These components work together to constrain the reasoning of the LLM, which ensures that the outputs remain consistent, and aligned with agronomic knowledge.

Figure 3 shows the prompt that guides SPADE. Each instruction in this prompt serves a distinct function in constraining and guiding SPADE's reasoning. First, the prompt begins with a clear task definition, explicitly specifying that the model must detect anomalies and summarize irrigation events from the given soil moisture data. Such explicit task specification narrows the scope of reasoning and improves consistency and reliability of the outputs (White et al., 2023). The input format is also specified, describing that the chronological table of timestamps and soil moisture values is provided to the model as a text sequence rather than as numeric data. This prompt ensures that the temporal order is interpreted correctly. The anomaly rules provide criteria for patterns that deviate from normal behavior, including extreme values, sharp spikes or dips, persistent level shifts, and missing data. In particular, unlike existing methods that focus solely on anomaly detection, SPADE simultaneously identifies both anomalies and irrigation patterns. To prevent these two types of events from being confused with each other, SPADE explicitly re-examines any suspicious patterns that occur near irrigation events using a zero-shot chain-of-thought (CoT) reasoning process (as specified in Anomaly Rule 7). Zero-shot CoT is known to improve reasoning by enabling LLMs to generate step-by-step solutions in a zero-shot setting (Kojima et al., 2022). By leveraging this capability, SPADE ensures that patterns overlapping with irrigation events are reassessed systematically, reducing the risk of misclassifying valid irrigation as anomalies. The domain rules guide the identification of valid irrigation events by encoding agronomic knowledge. They begin by defining commonly used expert concepts such as pre-spike baselines, plateaus, and declines, and establish thresholds for net gain. These rules also capture the intuitive decision-making process of domain experts by describing in words the typical behavior of normal irrigation patterns, which can convey nuances that are difficult to formalize mathematically (see Domain Rule 2). After analyzing the data by applying both sets of rules, SPADE generates a structured report that follows the predefined response format in Figure 3. This standardized output structure improves reproducibility and makes the results easier to interpret and understand. The generated report includes two main components: an anomaly report and an irrigation report. The anomaly report includes a binary indicator specifying whether any anomaly exists in the time series. If any anomaly is detected, SPADE provides a list of specific timestamps or time ranges indicating when the anomaly occurs, and classifies it into a predefined category. It also generates a concise textual explanation that shows the reasoning behind the classification. The irrigation report includes the following items: (i) a summary statement reporting the number of valid irrigation or rainfall events detected in the given segment; (ii) an event field listing the exact timestamps of each identified irrigation or rainfall event. For each event, the irrigation effectiveness is evaluated by calculating the net gain, which is computed as the difference between the plateau value and the pre-spike baseline moisture value; (iii) a key irrigation event section specifying the event with the highest net gain, including its timestamp; and (iv) a final net gain field representing the total net gain accumulated from all valid events in the segment. This information could provide practical insights for farmers to evaluate irrigation effectiveness and support decision-making for future irrigation scheduling.

### *2.3. Evaluation metrics*

We evaluate the performance of SPADE on 100 time-series samples for two tasks: irrigation pattern detection and anomaly detection. To evaluate its effectiveness, we compared SPADE against FlagIT (Dorigo et al., 2013) that is an existing zero-shot anomaly detection method specifically designed for soil moisture time-series data. FlagIT detects anomalies by applying threshold-based rules on absolute values and their temporal derivatives. Since SPADE is a zero-shot method that does not require training data, our comparison focused solely on FlagIT.

For evaluation, domain experts with professional experience in irrigation and soil moisture analysis manually reviewed the results of each method. Given the graph of a soil moisture time-series segment and the corresponding results generated by each method, experts were first asked to report the number of irrigation events and evaluate whether the method had correctly identified them. They then verified whether the key irrigation event had been accurately recognized and calculated the final net gain. Similarly, experts reported the total number of anomalies present in the data. If anomalies were detected, they assessed whether each method had successfully identified them and whether the corresponding anomaly types and explanations were appropriate. The expert annotations were used as ground truth for performance evaluation.

Based on these evaluations, both irrigation detection and anomaly detection tasks are assessed using standard classification metrics, including precision, recall, and F1-score (F1). These metrics are calculated as follows:

$$precision = \frac{TP}{FP + TP}. \tag{1}$$

$$recall = \frac{TP}{FN + TP}. \tag{2}$$

$$F1 = \frac{2 \times precision \times recall}{precision + recall}. \tag{3}$$

In the irrigation detection task, true positives (TP) refer to irrigation events correctly identified, false positives (FP) correspond to non-irrigation events incorrectly reported as irrigation, and false negatives (FN) indicate actual irrigation events that are not detected. In contrast, for the anomaly detection task, TP denotes anomalies accurately detected, FP indicates normal

## Task
You are an expert in soil moisture data analysis. Given the following time-series soil moisture data, your task is to: (1) Detect any anomalies in the data following anomaly rules. (2) Identify and report irrigation/rainfall events following domain rules.

## Input Format
You are provided with time-series soil moisture data in a table format. Each row contains a Timestamp column (format: YYYY-MM-DD HH:MM:SS)and a Moisture Value (%) column. The rows are sorted in chronological order.

## Anomaly Rules
1. An anomaly is any patterns that does not follow the normal irrigation/rainfall pattern.
2. It is very important if you are not sure whether a pattern is an anomaly, do not mark as anomaly and do not include in anomaly report.
3. Soil moisture values outside 5–60% are considered anomalies.
4. anomaly type should be one of the following:-
   - SingleSpike: Sharp rise in moisture not followed by a gradual decrease, but by a sharp drop to pre-rise level within 1 data point.
   - SingleDip: Sharp drop in moisture not followed by a gradual increase, but by a sharp rise to pre-drop level within 1 data point.
   - MultipleSpikes: Multiple brief rises with immediate decreases to pre-rise level.
   - MultipleDips: Multiple brief drops with immediate increases to pre-drop value.
   - Persistent Level Shift Up: The data shifts to a higher value and maintains that level consistently and constantly, do not return to the pre-surge level.
   - Persistent Level Shift Down: The data shifts to a lower value and maintains that level consistently and constantly, do not return to the pre-drop level.
   - Transient Level Shift Up: The data temporarily shifts to a higher value and then returns to the pre-rise level, anomaly maintains for at least 5 data points.
   - Transient Level Shift Down: The data temporarily shifts to a lower value and then returns to the pre-drop level, anomaly maintains for at least 5 data points.
   - MissingValue: No data available for 30 consecutive days or more.
   - No: No anomaly detected.
5. If more than five anomaly points are detected, do not list them individually. Instead, summarize using a time range that includes all anomaly points in anomaly report. This must be strictly followed. For spike and dip anomalies, only sharp rise and sharp drop are anomalies.
6. If more than 10 normal irrigation events are counted, those irrigation events are actually anomaly due to sensor malfunction, so do not include in irrigation report and include in anomaly report. This rule is the second-highest priority rule.
7. Reassess the anomalies thinking step by step and output your final analysis observing given response format.
8. If the final anomaly type is 'no', it is not an anomaly. Please revise the anomaly report.

## Domain Rules
1. Terminology
   - **Initial surge**: Sharp rise at the start of the event.
   - **Plateau**: Moisture 6 hours after the spike.
   - **Decline**: The decrease in moisture starting immediately after the peak.
   - **Net gain**: (Plateau – pre-surge moisture value).
   - **Final net gain**: The sum of net gains from all valid events.
2. A normal irrigation or rainfall event is characterized by (i) a sharp or gradual increase followed by a gradual decrease over time or (ii) a sharp or gradual increase followed by an initial sudden drop, then transitioning into a gradual decline. Normal irrigation can have a sharp increase higher than 8 percentage points within 2 hours, but the point that differs from anomaly is gradual decrease that is represented by a step-wise or curved descent in the moisture value. High sharpness and magnitude and graduality of the rise do not mean that it is not typical irrigation/rainfall events, so do not mark as anomaly if the data is following these patterns.
3. This rule should be strictly observed: if the net gain is smaller than 3, it is not a normal irrigation event, do not include in irrigation report.
4. This rule should be strictly observed: normal irrigation event does not include anomaly. If it is part of the anomaly event, it is not normal irrigation event, so do not include in irrigation report.
5. This is the highest-priority rule: even if a pattern resembles a normal irrigation event, any timestamps within 2 hours before or after of an anomaly are not as a normal irrigation event. A timestamp is considered to be within 2 hours of another timestamp if it is either within 2 hours before or within 2 hours after.
6. If multiple patterns resembling a normal irrigation event are observed (even if each surge is below or above the threshold), they are all normal irrigation events.
7. No other irrigation event occurs within 1 hour, and only the one with the larger increase is included in irrigation report.
8. Reassess any valid irrigation event is within 2 hours before or after from anomaly event. If it falls within, it is not a valid irrigation event, exclude it from the irrigation report.
9. Count the number of each valid irrigation event at the end and write it down in moisture report.

## Response Format
Respond in this strict JSON format with no additional text or commentary or explanation:
{ "anomalies": { "is anomaly": "true" or "false", "anomalies": "[timestamp1, timestamp2, ...]" or "time range string", "reason for anomaly type": "1 sentence reason within 20 word" or "no", "anomaly type": "classification of main anomaly" or "no" }, "irrigation report": { "summary": "In the shown data, there were xx valid irrigation or rainfall events.", "event": {"time": "[One highest peak increase timestamp in each irrigation event]"}, "key event": {"time":"[timestamp of event with the highest net gain]", "pre spike": "moisture value before the spike", "plateau": "moisture value 6 hours after spike", "net gain": "plateau - pre surge"}, "final net gain": "total net gain from all irrigation events" } events incorrectly flagged as anomalous, and FN represents true anomalies that the methods fail to detect. In addition to these classification metrics, the mean squared error (MSE) is computed to evaluate the accuracy of SPADE's net gain estimation. Specifically, we measure the difference between the final net gain value reported by SPADE and the expert-annotated final net gain value using the following formula:

$$MSE = \frac{1}{n}\sum_{i=1}^{n}(y_i - \hat{y}_i)^2, \tag{4}$$

where $n$ is the total number of samples, $y_i$ and $\hat{y}_i$ denote the annotated and predicted final net gain values, respectively. For anomaly detection, we also assess the accuracy of the anomaly type classification provided by each method.

Figure 3: Prompt for Soil Moisture Data Analysis with SPADE.

## 3. Result

### *3.1. Case Study*

As shown in Table 1, our proposed SPADE model showed promising performances in both anomaly detection and irrigation detection tasks. In the anomaly detection task, SPADE achieved a substantially higher recall and F1 score compared to FlagIT, while maintaining a competitive precision. FlagIT demonstrated a high precision, but showed a low recall. This result suggests that FlagIT is effective in detecting pronounced anomalies predominantly involving extreme deviations but fails to capture more subtle anomalous patterns. Furthermore, the anomaly type classification accuracy of FlagIT was low. We observed that it could reliably detect only SingleSpike anomalies with large deviations, while failing to distinguish other types. This limitation might arise from its threshold- and derivative-based detection strategy, which is susceptible to missing anomalies that do not exceed predefined thresholds. Notably, SPADE also outperformed FlagIT in anomaly type classification accuracy, which demonstrates its ability not only to detect anomalies but also to correctly identify their types. In particular, our dataset contained all eight types of anomalies, and SPADE was able to correctly classify them across these diverse anomaly types. For irrigation detection, SPADE achieved high precision, recall, and F1 scores under variability in normal irrigation patterns. MSE for net gain estimation was 5.78, suggesting SPADE could assist farmers in evaluating irrigation effectiveness.

As shown in Figure 4, for each soil moisture time series exhibiting multiple irrigation events, SPADE generated reports that accurately identified all valid irrigation events (highlighted in light blue) across all examples. Despite being a language model, it accurately identified each key event with the highest net gain and produced a reasonably accurate estimates of the total net gains across all examples. Furthermore, it did not report any anomalies, demonstrating its ability to distinguish irrigation patterns from anomalies. These examples suggests that SPADE can adapt its analysis to different normal irrigation patterns, producing structured reports even when the patterns vary considerably.

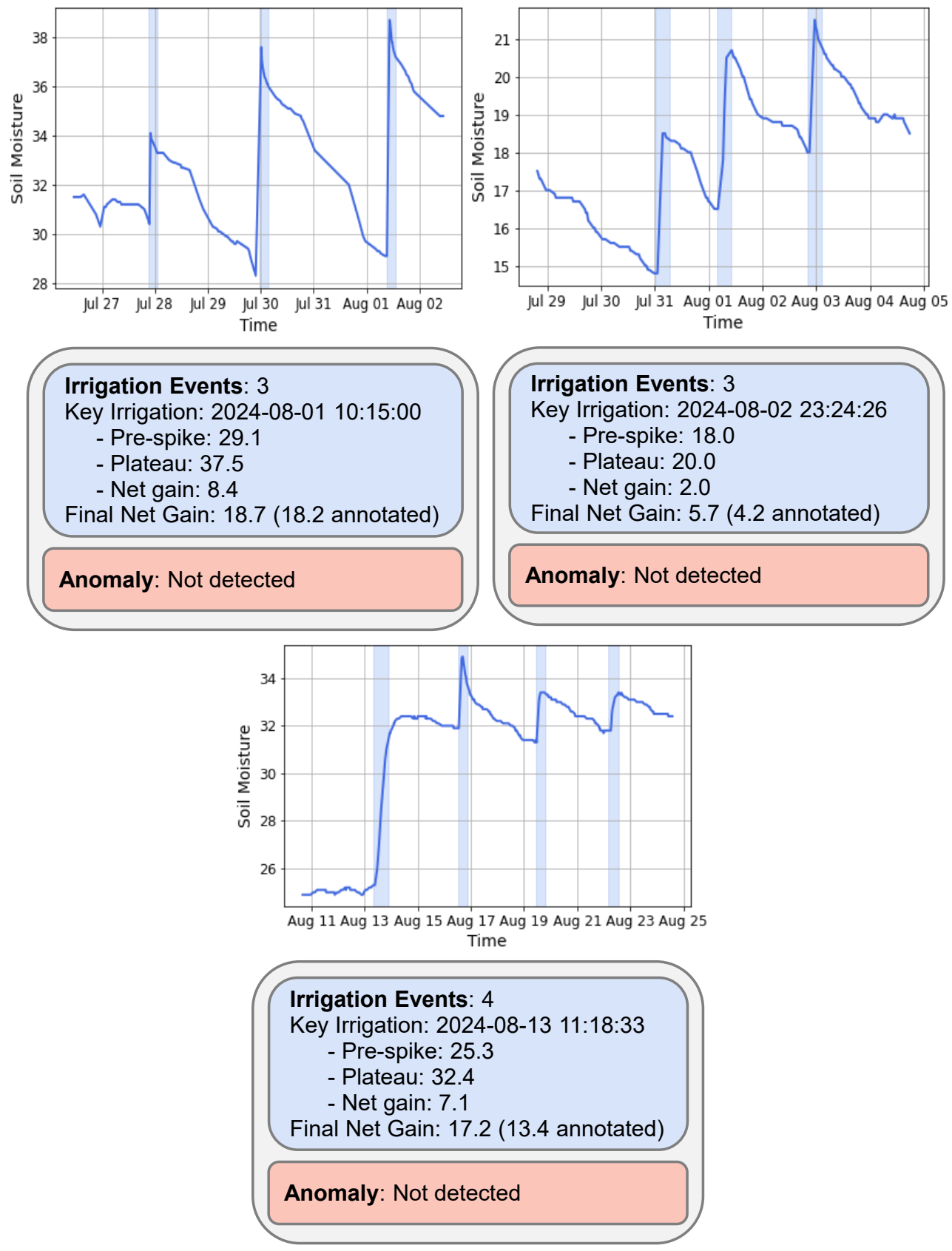

Figure 4: Examples of SPADE-generated reports paired with corresponding weekly soil moisture data with various normal irrigation patterns. The input of soil moisture data to SPADE consisted of numeric soil moisture values over time, provided as strings but shown here as line graphs for interpretability. The x-axis represents time, and the y-axis indicates volumetric soil moisture content (%).

In contrast, Figure 5 shows the soil moisture time-series plots containing different type of anomalies, while the corresponding SPADE-generated reports are displayed below. As shown in Figure 5, SPADE correctly identified all types of anomalies in the examples and accurately pinpointed the exact time of occurrence. It also generated concise, interpretable explanations for why these patterns were classified as anomalies, drawing on its reasoning capabilities as a language model. As illustrated in the bottom right example in Figure 5, SPADE was also able to detect valid irrigation events that occurred alongside the anomalies.

### *3.2. Representative Examples of Prompt Instruction Effectiveness*

We validated the impact of each instruction in the prompt by presenting two representative examples focused on Domain Rule 2 and Anomaly Rule 7. Domain Rule 2 specifies the general behavior of normal irrigation or rainfall events (see Figure 3). This rule provides context on unique characteristics of normal irrigation patterns in soil moisture data. Anomaly Rule 7 leverages a zero-shot CoT reasoning (see Figure 3) (Wang et al., 2023). To examine the effect of each rule, we generated reports with and without the corresponding rule. As illustrated in Figure 6, the top plot shows a soil moisture time series with one irrigation event verified by domain experts, highlighted in light blue. The two report panels below the graph compare SPADE's outputs with and without Domain Rule 2 given the time series. The left report (w/o Domain Rule 2) was generated without the rule, while the right report includes it. Without Domain Rule 2, the actual irrigation event was mistakenly identified as a SingleSpike anomaly (left panel), whereas SPADE correctly identified it as a valid irrigation event when the rule was included (right panel). This example demonstrates that Domain Rule 2 provides the necessary context on irrigation patterns in soil moisture data. Similarly, Figure 7 presents a case demonstrating the impact of Anomaly Rule 7. Given a time series with one valid irrigation event (top plot), SPADE correctly identified it by reassessing the pattern through step-by-step reasoning with Anomaly Rule 7. However, without the rule, the same event was mistakenly detected as an anomaly. This result highlights the effectiveness of Anomaly Rule 7 in enabling zero-shot CoT reasoning, which helps SPADE to detect anomalies more accurately.

Table 1: Performance of anomaly detection and irrigation detection with 100 segmented time series. Higher values for each metric are highlighted in bold.

| | Anomaly Detection | | | | Irrigation Detection | | | |
|---|---|---|---|---|---|---|---|---|
| | Precision | Recall | F1 | Anomaly Type Classification Accuracy | Precision | Recall | F1 | Net Gain MSE (%) |
| FlagIT (Dorigo et al., 2013) | **0.90** | 0.31 | 0.51 | 0.36 | - | - | - | - |
| SPADE (proposed) | 0.85 | **0.97** | **0.91** | **0.96** | 0.97 | 0.91 | 0.94 | 5.78 |

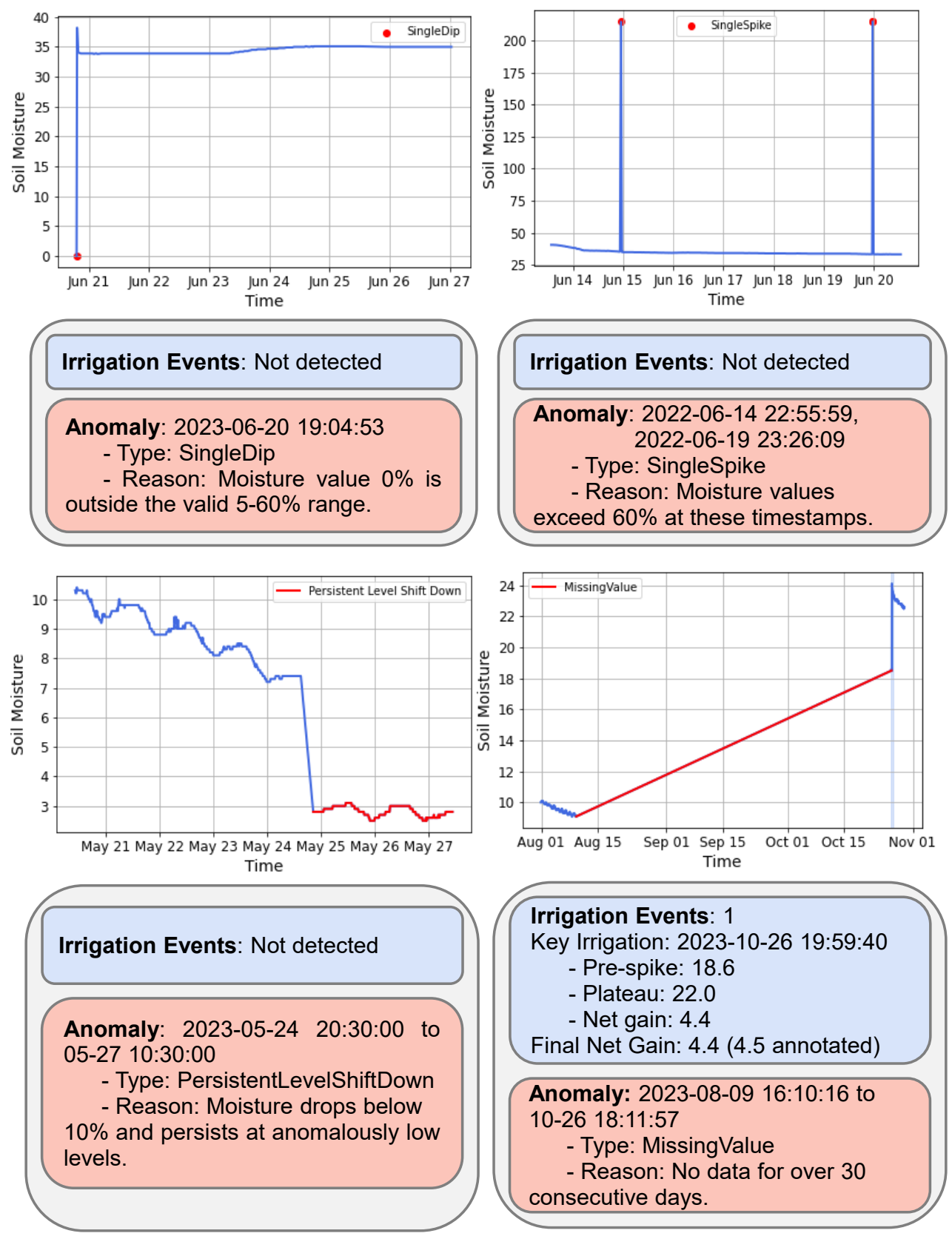

Figure 5: Examples of SPADE-generated reports paired with corresponding weekly soil moisture data containing different types of anomalies. The input of soil moisture data to SPADE consisted of numeric soil moisture values over time, provided as strings but shown here as line graphs for interpretability. The x-axis represents time, and the y-axis indicates volumetric soil moisture content (%).

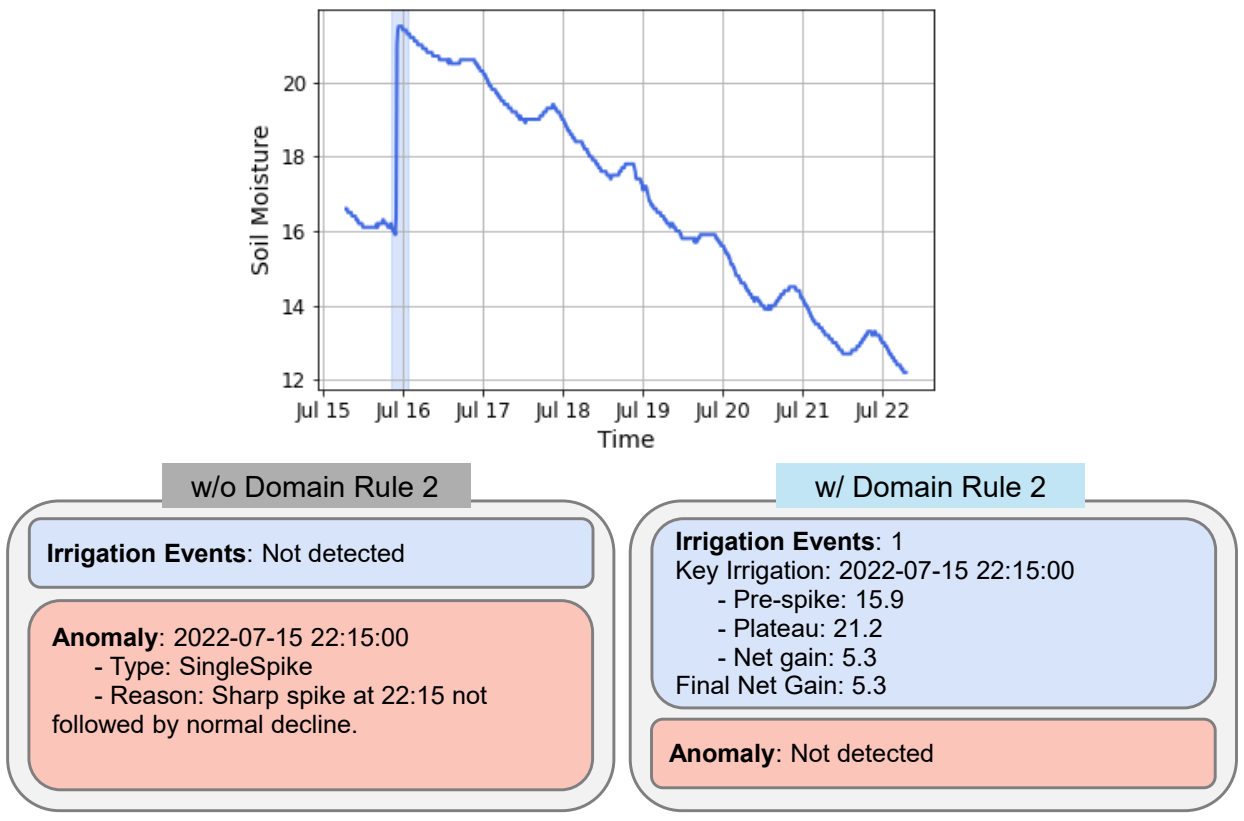

Figure 6: Comparison of SPADE-generated reports with (right panel) and without (left panel) Domain Rule 2 for the same soil moisture time series. The input of soil moisture data to SPADE consisted of numeric soil moisture values over time, provided as strings but shown here as a line graph for interpretability. The x-axis represents time, and the y-axis indicates volumetric soil moisture content (%).

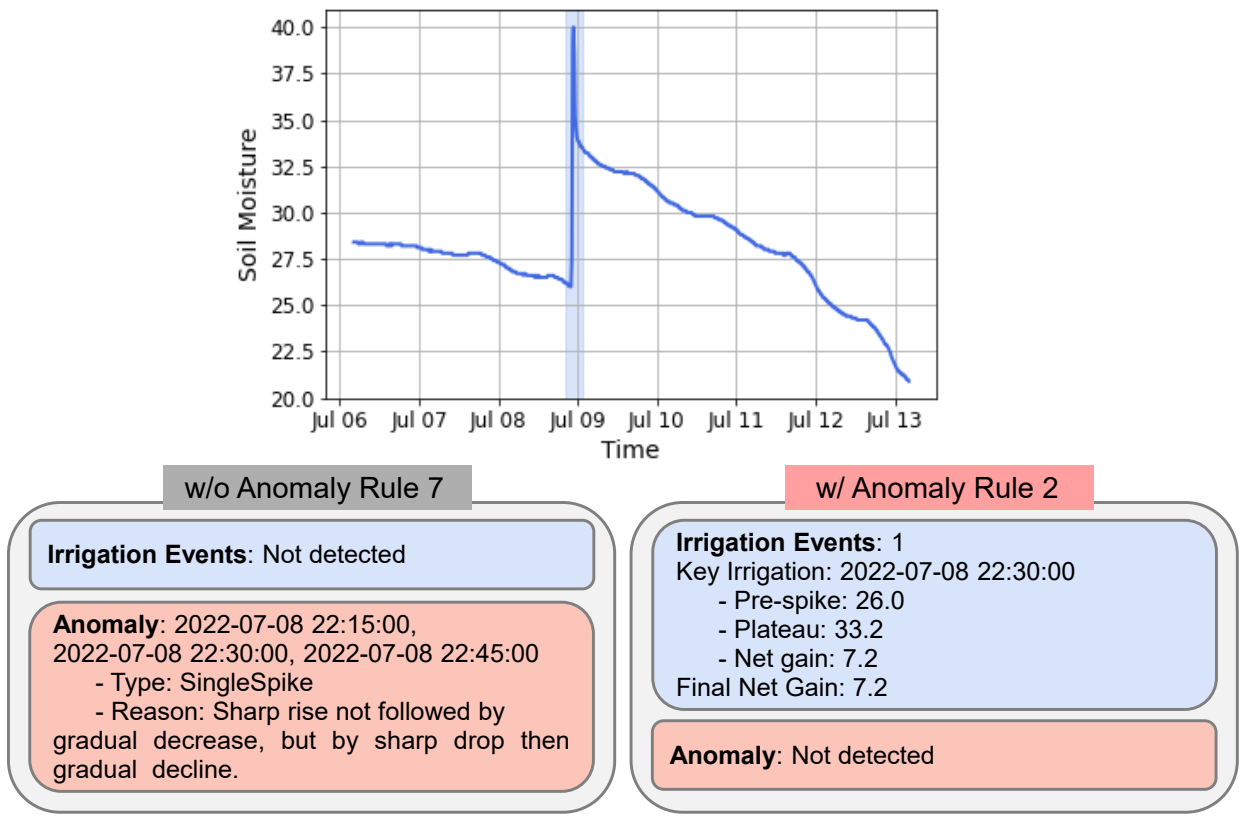

Figure 7: Comparison of SPADE-generated reports with (right panel) and without (left panel) Anomaly Rule 7 for the same soil moisture time series. The input of soil moisture data to SPADE consisted of numeric soil moisture values over time, provided as strings but shown here as a line graph for interpretability. The x-axis represents time, and the y-axis indicates volumetric soil moisture content (%).

### 3.3. *Cost Evaluation*

To assess the feasibility of deploying SPADE in real-world soil moisture sensor monitoring systems, we conducted a cost evaluation of SPADE. SPADE employed ChatGPT 4.1 released on April 14. The input token cost for ChatGPT 4.1 is $2.00 per one million tokens. On average, a single weekly farm dataset was converted into approximately 3,000 tokens, resulting in an input cost of roughly $0.006 per dataset. The output token cost is $8.00 per one million tokens. The generated reports typically consist of around 150 tokens, depending on the number of irrigation events and detected anomalies, which corresponds to an output cost of approximately $0.0012. Consequently, the total cost of generating one report per weekly farm input is estimated to be around $0.0072.

## 4. Discussion

This study demonstrates the potential of LLMs as decision-support tools in agriculture by enabling them to reason about soil moisture dynamics and irrigation practices in context. SPADE shows that LLMs can interpret complex time-series patterns from sensor data and generate structured, human-readable reports that summarize irrigation events and anomalies. This interpretability is particularly valuable in agricultural settings, where decision-making often depends on nuanced temporal trends that traditional threshold-based or black-box models struggle to explain.

One key advantage of SPADE is its ability to integrate qualitative domain knowledge with data-driven reasoning. By leveraging zero-shot prompting and CoT reasoning, SPADE can distinguish irrigation events from anomalies, identify subtle irrigation anomalies, and provide insights that are directly relevant to farmers. Furthermore, its zero-shot capability allows the model to operate without task-specific annotations or fine-tuning, which enhances adaptability across farms and sensor configurations.

Nonetheless, the practical deployment of LLMs continues to pose several challenges. LLMs are known to be sensitive to prompt phrasing and might produce variable outputs across runs, making consistent reporting difficult (Gallifant et al., 2025). Although we did not observe hallucinations in our experiments, we are aware that it is a well-known limitation of LLMs (Huang et al., 2025). In our case, the structured nature of the sensor data and prompt design, such as the use of zero-shot CoT, likely reduced this risk. Nevertheless, future extensions could incorporate rule-based output validation to further improve reliability (White et al., 2023; Jiang et al., 2025). Additionally, the computational cost of running LLMs at scale could be a barrier to real-time deployment and may require optimization for practical use.

SPADE has limitations that point to promising directions for future research. First, our analysis focused on weekly segments rather than full seasonal records, which might limit the ability to capture longer-term irrigation dynamics. As computational resources permit, future work could extend the analysis to full-season data. Second, we used only one randomly selected depth per farm, whereas incorporating the full multi-depth structure could yield more robust insights. Finally, while SPADE currently focuses on interpreting patterns and detecting anomalies, future versions could explore whether it can also provide irrigation recommendations based on its analysis, which can offer more actionable guidance for precision agriculture.

## 5. Conclusion

In this paper, we proposed SPADE, a novel LLM-based framework for analyzing soil moisture time-series data and detecting irrigation events and anomalies. To the best of our knowledge, this is the first study to apply a large language model to the analysis of field-collected soil moisture data. SPADE translates weekly volumetric soil moisture data into structured reports that summarize irrigation patterns and detect potential anomalies without requiring training data or additional training. By providing timely anomaly alerts and interpretable insights into irrigation effectiveness, SPADE could support farmers' decision-making processes and improve irrigation management.

## Declaration of competing interest

The authors declare that they have no known competing financial interests or personal relationships that could have appeared to influence the work reported in this paper.

## Acknowledgements

This work was supported by the USDA National Institute of Food and Agriculture (NIFA) under Award No. 2024-67021-41534. We would also like to express our sincere appreciation to the collaborating growers and interns whose efforts in data collection greatly contributed to this research.

## Data Availability

The research plot data supporting the findings of this study are available upon request. Due to privacy agreements, data from collaborating growers are not publicly available.

## References

Achiam, J., Adler, S., Agarwal, S., Ahmad, L., Akkaya, I., Aleman, F.L., Almeida, D., Altenschmidt, J., Altman, S., Anadkat, S., et al., 2023. Gpt-4 technical report. arXiv preprint arXiv:2303.08774 .

Alnegheimish, S., Nguyen, L., Berti-Equille, L., Veeramachaneni, K., 2024. Large language models can be zero-shot anomaly detectors for time series? arXiv preprint arXiv:2405.14755 .

Bandaru, L., Irigireddy, B.C., Davis, B., et al., 2024. Deepqc: A deep learning system for automatic quality control of in-situ soil moisture sensor time series data. Smart Agricultural Technology 8, 100514.

Catalano, C., Paiano, L., Calabrese, F., Cataldo, M., Mancarella, L., Tommasi, F., 2022. Anomaly detection in smart agriculture systems. Computers in Industry 143, 103750.

Dorigo, W., Xaver, A., Vreugdenhil, M., Gruber, A., Hegyiova, A., Sanchis-Dufau, A.D., Zamojski, D., Cordes, C., Wagner, W., Drusch, M., 2013. Global automated quality control of in situ soil moisture data from the international soil moisture network. Vadose Zone Journal 12, vzj2012–0097.

Farooq, A., Farooq, N., Akbar, H., Hassan, Z.U., Gheewala, S.H., 2023. A critical review of climate change impact at a global scale on cereal crop production. Agronomy 13, 162.

Gallifant, J., Afshar, M., Ameen, S., Aphinyanaphongs, Y., Chen, S., Cacciamani, G., Demner-Fushman, D., Dligach, D., Daneshjou, R., Fernandes, C., et al., 2025. The tripod-llm reporting guideline for studies using large language models. Nature medicine 31, 60–69.

Huang, L., Yu, W., Ma, W., Zhong, W., Feng, Z., Wang, H., Chen, Q., Peng, W., Feng, X., Qin, B., et al., 2025. A survey on hallucination in large language models: Principles, taxonomy, challenges, and open questions. ACM Transactions on Information Systems 43, 1–55.

Jalilvand, E., Tajrishy, M., Hashemi, S.A.G.Z., Brocca, L., 2019. Quantification of irrigation water using remote sensing of soil moisture in a semi-arid region. Remote Sensing of Environment 231, 111226.

Jiang, Y., Chen, J., Yang, D., Li, M., Wang, S., Wu, T., Li, K., Zhang, L., 2025. Comt: Chain-of-medical-thought reduces hallucination in medical report generation, in: ICASSP 2025-2025 IEEE International Conference on Acoustics, Speech and Signal Processing (ICASSP), IEEE. pp. 1–5.

Kojima, T., Gu, S.S., Reid, M., Matsuo, Y., Iwasawa, Y., 2022. Large language models are zero-shot reasoners. Advances in neural information processing systems 35, 22199–22213.

Liu, C., He, S., Zhou, Q., Li, S., Meng, W., 2024a. Large language model guided knowledge distillation for time series anomaly detection. arXiv preprint arXiv:2401.15123 .

Liu, J., Zhang, C., Qian, J., Ma, M., Qin, S., Bansal, C., Lin, Q., Rajmohan, S., Zhang, D., 2024b. Large language models can deliver accurate and interpretable time series anomaly detection. arXiv preprint arXiv:2405.15370 .

Mittelbach, H., Lehner, I., Seneviratne, S.I., 2012. Comparison of four soil moisture sensor types under field conditions in switzerland. Journal of Hydrology 430, 39–49.

Muleke, A., Harrison, M.T., Eisner, R., de Voil, P., Yanotti, M., Liu, K., Monjardino, M., Yin, X., Wang, W., Nie, J., et al., 2023. Sustainable intensification with irrigation raises farm profit despite climate emergency. Plants, People, Planet 5, 368–385.

OpenAI, 2025. Introducing gpt-4.1 in the api. `https://openai.com/index/gpt-4-1/`. Accessed: 2025-07-26.

Russell-Gilbert, A., Sommers, A., Thompson, A., Cummins, L., Mittal, S., Rahimi, S., Seale, M., Jaboure, J., Arnold, T., Church, J., 2024. Aad-llm: Adaptive anomaly detection using large language models, in: 2024 IEEE International Conference on Big Data (BigData), IEEE. pp. 4194–4203.

Wang, L., Xu, W., Lan, Y., Hu, Z., Lan, Y., Lee, R.K.W., Lim, E.P., 2023. Plan-and-solve prompting: Improving zero-shot chain-of-thought reasoning by large language models. arXiv preprint arXiv:2305.04091 .

White, J., Fu, Q., Hays, S., Sandborn, M., Olea, C., Gilbert, H., Elnashar, A., Spencer-Smith, J., Schmidt, D.C., 2023. A prompt pattern catalog to enhance prompt engineering with chatgpt. arXiv preprint arXiv:2302.11382 .